\def\year{2021}\relax
\documentclass[letterpaper]{article} 
\usepackage{aaai21}  
\usepackage{times}  
\usepackage{helvet} 
\usepackage{courier}  
\usepackage[hyphens]{url}  
\usepackage{graphicx} 
\usepackage{natbib}  
\usepackage{caption} 
\usepackage[switch]{lineno}  %
\usepackage{graphicx}
\usepackage{amsmath}
\usepackage{amssymb}
\usepackage{color}
\usepackage{bm}
\usepackage{amsthm} 
\usepackage{algorithm}
\usepackage{algorithmic}
\newtheorem{definition}{Definition}
\usepackage{caption}
\usepackage{stmaryrd}
\usepackage{MnSymbol}
\usepackage{url}

\newcommand{\bmx}{\bm{x}}

\title{Anomaly Attribution with Likelihood Compensation}

\author{%
	Tsuyoshi Id\'{e}, \ Amit~Dhurandhar, \ Ji\v{r}\'\i~Navr\'atil, \ Moninder~Singh,  Naoki~Abe \\
}
\affiliations {
	IBM Research, T. J. Watson Research Center\\  \{tide, adhuran, jiri, moninder, nabe\}@us.ibm.com
}

\begin{document}

\maketitle

\begin{abstract}
This paper addresses the task of explaining anomalous predictions of a black-box regression model. When using a black-box model, such as one to predict building energy consumption from many sensor measurements, we often have a situation where some observed samples may significantly deviate from their prediction. It may be due to a sub-optimal black-box model, or simply because those samples are outliers. In either case, one would ideally want to compute a ``responsibility score'' indicative of the extent to which an input variable is responsible for the anomalous output. In this work, we formalize this task as a statistical inverse problem: Given model deviation from the expected value, infer the responsibility score of each of the input variables. We propose a new method called likelihood compensation (LC), which is founded on the likelihood principle and computes a correction to each input variable. To the best of our knowledge, this is the first principled framework that computes a responsibility score for real valued anomalous model deviations. We apply our approach to a real-world building energy prediction task and confirm its utility based on expert feedback.
%
\end{abstract}


\section{Introduction}\label{sec:Introduction}

With the rapid development of Internet-of-Things technologies, anomaly detection has played a critical role in modern industrial applications of artificial intelligence (AI). One of the recent technology trends is to create a ``digital twin'' using a highly flexible machine learning model, typically deep neural networks, for monitoring the health of the production system~\cite{tao2018digital}. However, the more representational power the model has, the more difficult it is to understand its behavior. In particular, explaining deviations between predictions and true/expected measurements is one of the main pain points. A large deviation from the truth may be due to sub-optimal model training, or simply because the observed samples are outliers. If the model is black-box and the training dataset is not available, it is hard to determine which of these two situations have occurred. Nonetheless, we would still want to provide information to help end-users' in their decision making. 

As such, in this paper we propose a method that can compute a 
``responsibility score'' for each variable of a given input. We aver to this task as \textit{anomaly attribution}. Specifically, we are concerned with model-agnostic anomaly attribution for black-box \textit{regression} models, where we want to explain the deviation between the models prediction and the true/expected output in as concise a manner as possible. As a concrete example, consider the scenario of monitoring building energy consumption as the target variable $y$ (see Section~\ref{sec:experiments} for the detail). The input to the model is a multi-variate sensor measurement $\bm{x}$ that is typically \textit{real-valued} and \textit{noisy}. Under the assumption that the model is black-box and the training data are not available, our goal is to compute a numerical score for each of the input variables, quantifying the extent to which they are responsible for the judgment that a given test sample is anomalous.

Anomaly attribution has been studied typically as a sub-task of anomaly detection to date. For instance, in subspace-based anomaly detection, computing each variable's responsibility has been part of the standard procedure for years~\cite{Chandola09AnomalySurvey}. However, there is little work on how anomaly attribution can be done when the model is black-box and the training data set is not available. In the XAI (explainable AI) community, on the other hand, growing attention has been paid to ``post-hoc'' explanations of black-box prediction models. Examples of the techniques include feature subset selection, feature importance scoring, and sample importance scoring~\cite{Costabello2019AAAI_tutorial,molnar2019interpretable,samek2019explainable}. 
%
For anomaly attribution, there are at least two post-hoc explanation approaches that are potentially applicable: (1) those based on the expected conditional deviation, best known as the Shapley value, which was first introduced to the machine learning community by \v{S}trumbelj and Kononenko~\shortcite{kononenko2010efficient}, and (2) those based on local linear models, best known under the name of LIME (Local Interpretable Model-agnostic Explanations)~\cite{ribeiro2016should}. In spite of their popularity, these two approaches may not be directly useful for anomaly attribution: Given a test sample $(\bm{x}^t,y^t)$, these methods may explain what the value of $f(\bm{x}^t)$ itself can be attributed to. However, \textit{what is more relevant is explaining the deviation of $f(\bm{x}^t)$ from the actual $y^t$.}

\begin{figure}[t]
\begin{center}
\includegraphics[trim={4.8cm 3.6cm 5.2cm 3.3cm},clip,width=8.3cm]{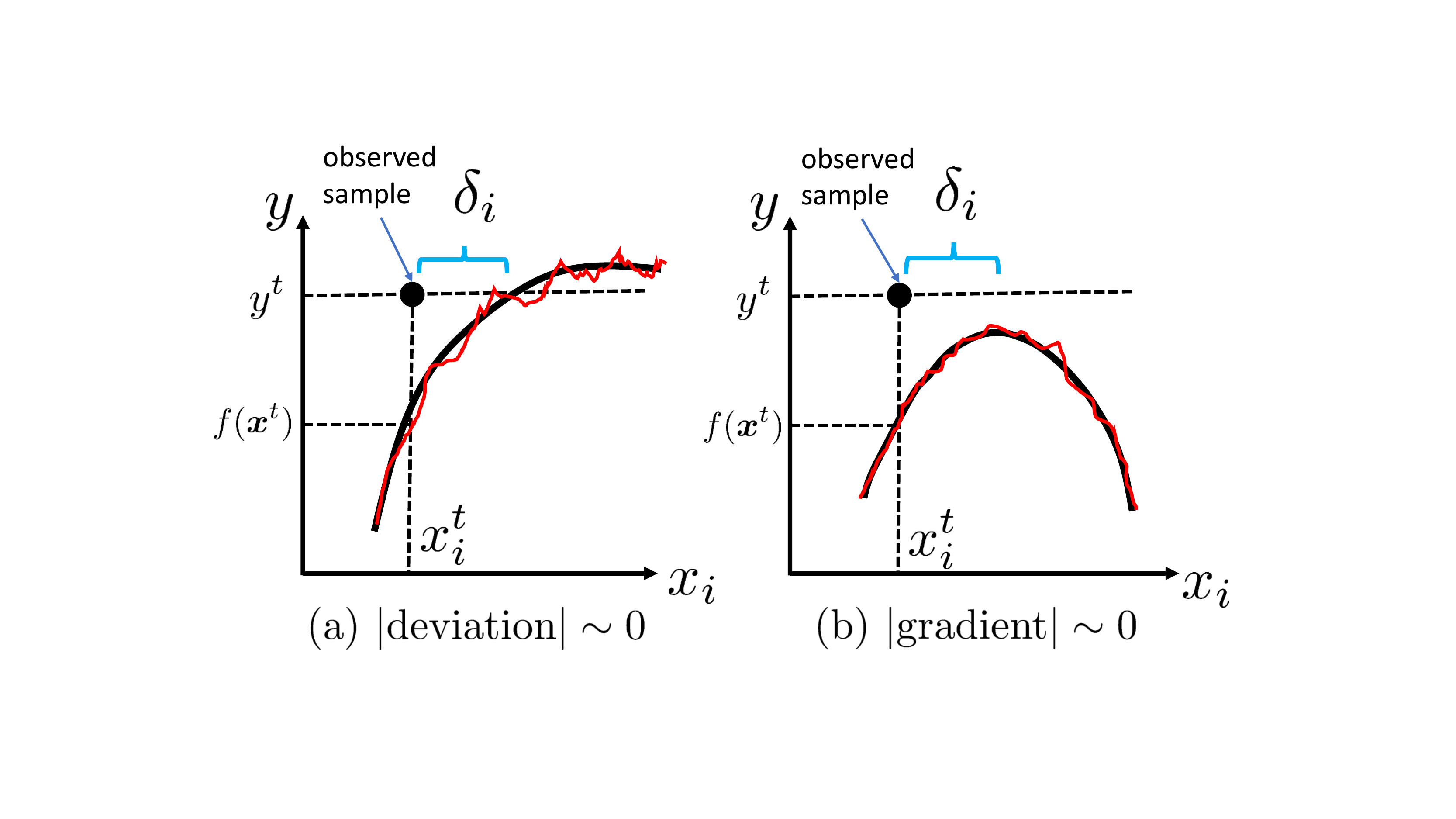}
\end{center}\vspace{-0.3cm}
\caption{Illustration of the likelihood compensation along the $i$-th axis ($\delta_i$). For a given test sample $(y^t,\bm{x}^t)$, LC seeks a perturbation that achieves the best possible fit with the black-box regression model $y=f(\bm{x})$, which could be non-smooth (the red curves). For more details please refer to Section \ref{Sec:MOC}. }
\label{fig:OCillustration.pdf}
\end{figure}

To address this limitation, we propose \textit{likelihood compensation} (LC), a new local anomaly attribution approach for black-box regression models. We formalize the task of anomaly attribution as a statistical \textit{inverse problem} that infers a perturbation to the test input $\bmx^t$ from the deviation $y^t - f(\bmx^t)$, conversely to the forward problem that computes the deviation (or its variants) from  $(\bm{x}^t,y^t)$. As illustrated in Fig.~\ref{fig:OCillustration.pdf}, LC can be viewed intuitively as the ``deviation measured horizontally'' if certain conditions are met. This admits direct interpretation as it suggests an action that might be taken to bring back the outlying sample to normalcy. Importantly, LC does not use any problem-specific assumptions but is built upon the maximum likelihood principle, the basic principle in statistical machine learning. To the best of our knowledge, this is the first principled framework for model-agnostic anomaly attribution in the regression setting. 

\section{Related Work}

Although the machine learning community had not paid much attention to explainability of AI (XAI) in the black-box setting until recently, the last few years has seen a surge of interest in XAI research. For general background, Gade et al.~\shortcite{gade2020explainable} provides a useful summary of major research issues in XAI for industries. In a more specific context of inddustrial anomaly detection, Langone et al.~\shortcite{langone2020interpretable} and Amarasinghe et al.~\shortcite{amarasinghe2018toward} give a useful summary of practical requirements of XAI in the deep learning era. An extensive survey on various XAI methodologies is given in~\cite{Costabello2019AAAI_tutorial,molnar2019interpretable,samek2019explainable}.

So far most of the model-agnostic post-hoc explanation studies are designed for classification, often restricted to image classification. As discussed before, two approaches are potentially applicable to the task of anomaly attribution in the black-box regression setting, namely the Shapley value (SV)~\cite{kononenko2010efficient,vstrumbelj2014explaining,casalicchio2018visualizing} and the LIME~\cite{ribeiro2016should,ribeiro2018anchors}. The relationship between the two was discussed by Lundberg~et~al.~\shortcite{lundberg2017unified} assuming binary inputs. In the context of anomaly detection from noisy, real-valued data, two recent studies, Zhang~et~al.~\shortcite{zhang2019anomaly} and Giurgiu~et~al.~\shortcite{giurgiu2019additive}, proposed a method built on LIME and SV, respectively. While these belong to the earliest model-agnostic XAI studies for anomaly detection, they naturally inherit the limitations of the existing approaches mentioned in introduction. Recently, Lucic et al.~\shortcite{lucic2020does} proposed a LIME-based approach for identifying a variable-wise normal range, which although related is different from our formulation of the anomaly attribution problem. Zemicheal et al.~\shortcite{Zemicheal19COMPASS} proposed an SV-like feature scoring method in the context of outlier detection, however, this does not apply to the regression setting.

One of the main contributions of this work is the proposal of a generic XAI framework for input attribution built upon the likelihood principle. The method of integrated gradient~\cite{sundararajan2017axiomatic} is another generic input attribution approach applicable to the black-box setting. Sipple~\shortcite{SippleICML2020} recently applied it to anomaly detection and explanation. However, it is applicable to only the classification setting and, as pointed out by Sipple~\shortcite{SippleICML2020}, the need for the ``baseline input'' makes it less useful in practice. Layer-wise relevance propagation~\cite{bach2015pixel} is another well-known input attribution method and has been applied to real-world anomaly attribution tasks~\cite{amarasinghe2018toward}. However, it is deep-learning-specific and assumes we have white-box access to the model.

Another research thread relevant to our work revolves around the counterfactual approach, which focuses on what is missing in the model (or training data) rather than what exists. In the context of image classification, the idea of counterfactuals is naturally translated into perturbation-based explanation~\cite{fong2017interpretable}. Recent works~\cite{dhurandhar2018explanations,wachter2017counterfactual} proposed the idea of contrastive explanation, which attempts to find a perturbation best characterizing a classification instance such that the probability of choosing a different class supersedes the original prediction. Our approach is similar in spirit, but as mentioned above, is designed for regression and uses a very different objective function. Moreover, both of these methods~\cite{dhurandhar2018explanations,wachter2017counterfactual} require white-box access, while ours is a black-box approach. 

\section{Problem Setting}

As mentioned before, we focus on the task of anomaly attribution in the \textit{regression setting} rather than classification or unsupervised settings. Throughout the paper, the input variable $\bm{x}$ is assumed to be noisy, multivariate, and real-valued in general. Our task is formally stated as follows:
\begin{definition}[\textbf{Anomaly detection and attribution}]
 Given a black-box regression model $y=f(\bm{x})$ and a test data set $\mathcal{D}_\mathrm{test}$: (1) compute the score to represent the degree of anomaly of the prediction on $\mathcal{D}_\mathrm{test}$; (2) compute the responsibility score for each input variable for the prediction being anomalous.
\end{definition}
The black-box regression model is assumed to be deterministic with $y \in \mathbb{R}$ and $\bm{x} \in \mathbb{R}^M$, where $M\geq 2$ is the dimensionality of the input random variable $\bm{x}$. The functional form of $f(\cdot)$ and the dependency on model parameters are not available to us. The \textit{training data} on which the model was trained is \textit{not} available  either. The only interface to the model we are given is $\bm{x}$, which follows an unknown distribution $P(\bm{x})$. Queries to get the response $f(\bm{x})$ can be performed cheaply at any $\bm{x}$.

The test data set is denoted as $\mathcal{D}_\mathrm{test}= \{ (\bm{x}^t,y^t) \mid t=1,\ldots,N_\mathrm{test}\}$, where $t$ is the index for the $t$-th test sample and $N_\mathrm{test}$ is the number of test samples. When $N_\mathrm{test}=1$, the task may be called the \textit{outlier detection and attribution}.

\paragraph{Anomaly detection as forward problem}
Assume that, from the deterministic regression model, we can somehow obtain $p(y\mid \bm{x})$, a probability density over $y$ given the input signal $\bm{x}$ (see Sec.~\ref{subsec:getting_probabilistic_model} for a proposed approach to do this). The standard approach to quantifying the degree of anomaly is to use the negative log-likelihood of test samples. Under the i.i.d.~assumption, it can be written as
\begin{align}\label{eq:changeScoreDef}
a(\mathcal{D}_\mathrm{test}) = -
\frac{1}{
	N_\mathrm{test}
}\sum_{t \in \mathcal{D}_\mathrm{test}}\ln p(y^t \mid \bm{x}^t) ,
\end{align}
which is called the \textit{anomaly score} for $\mathcal{D}_\mathrm{test}$ (or the \textit{outlier score} for a single sample dataset). An anomaly is declared when $a(\mathcal{D}_\mathrm{test})$ exceeds a predefined threshold.

\paragraph{Anomaly attribution as inverse problem}
The above anomaly/outlier \textit{detection} formulation is standard. 
However, the anomaly/outlier \textit{attribution} is more challenging when the underlying model is black-box. This is in some sense an \textit{inverse problem}: The function $f(\bm{x})$ readily gives an estimate of $y$ from $\bm{x}$, but, in general, there is no obvious way to do the reverse in the \textit{multivariate} case. When an estimate $f(\bm{x}^t)$ looks `bad' in light of an observed $y^t$, what can we say about the contribution, or responsibility, of the input variables? Section~\ref{Sec:MOC} below gives our proposed answer to this question.

\paragraph{Notation}
We use boldface to denote vectors. The $i$-th dimension of a vector $\bm{\delta}$ is denoted as $\delta_i$. The $\ell_1$ and $\ell_2$ norms of a vector are denoted by $\| \cdot \|_1$ and $\| \cdot \|_2$, respectively, and are defined as $\| \bm{\delta} \|_1 \triangleq \sum_i | \delta_i|$ and $\| \bm{\delta} \|_2 \triangleq \sqrt{\sum_i  \delta_i^2}$. The sign function $\mathrm{sign}(\delta_i) $ is defined as being $1$ for $\delta_i>0$, and $-1$ for $\delta_i <0$.
For $\delta_i=0$, the function takes a value in $[-1,1]$. For a vector input, the definition applies element-wise, giving a vector of the same size as the input.

We distinguish between a random variable and its realizations with a superscript. For notational simplicity, we symbolically use $p(\cdot)$ to represent different probability distributions, whenever there is no confusion. For instance, $p(\bm{x})$ is used to represent the probability density of a random variable $\bm{x}$ while $p(y \mid \bm{x})$ is a different distribution of another random variable $y$ conditioned on $\bm{x}$. The Gaussian distribution of a random variable $y$ is denoted by $\mathcal{N}(y \mid \cdot, \cdot )$, where the first and the second arguments after the bar are the mean and the variance, respectively. 
The multivariate Gaussian distribution is defined in a similar way.

\section{The Method of Likelihood Compensation}\label{Sec:MOC}

This section presents the key idea of ``likelihood compensation'' as illustrated in Fig.~\ref{fig:OCillustration.pdf}. We start with a likelihood-based interpretation of LIME to highlight the idea.

\subsection{Improving Likelihood via Corrected Input}
\label{subsec:LIMEandOC}

For a given test sample $(y^t,\bm{x}^t)$, LIME minimizes the lasso objective to let the sparse regression estimation process select a subset of the variables. From a Bayesian perspective, it can be rewritten as a MAP (maximum a posteriori) problem:
\begin{gather}\nonumber
\mbox{LIME: }
\max_{\bm{\beta}}\left\langle
\ln \left\{ p(y \mid \bm{x}^t, \bm{\beta})
\ p(\bm{\beta})
\right\}\right\rangle_{\mathrm{vic}(\bm{x}^t)}
\\
\quad\mbox{subject to} \quad y=f(\bm{x}),
\end{gather}
where $\langle \cdot \rangle_{\mathrm{vic}(\bm{x}^t)}$ denotes the expectation over random samples generated from an assumed local distribution in the vicinity of $\bm{x}^t$. For $p$'s above, LIME uses the Gaussian observation model $p(y \mid \bm{x}, \bm{\beta}) =\mathcal{N}(y \mid \beta_0 + \bm{\beta}^\top\bm{x}, \sigma^2)$ and the Laplace prior $ p(\bm{\beta})\propto \exp\left( -  \nu \|\bm{\beta} \|_1 \right)$. Here $\sigma^2, \nu$ are hyperparameters. The regression coefficient $\bm{\beta}$ as well as the intercept $\beta_0$ captures the local linear structure of $f$ and is interpreted as the sensitivity of $f$ at $\bm{x}^t$.

From the viewpoint of actionability, however, the slope can be less useful than $\bm{x}$ itself, particularly for the purpose of outlier attribution. If $(\bm{x}^t,y^t)$ is an outlier far from the population, how can we expect to obtain actionable insights from the local slope? Another issue is that $y^t$ plays no role in this formulation. Notice the constraint of maximization: LIME amounts to assuming that the model is always right and is not sensitive to the question of whether $(y^t,\bm{x}^t)$ is an outlier or not.

Keeping this in mind, we propose to introduce a directly interpretable parameter $\bm{\delta}$ as a correction term to $\bm{x}$, rather than the slope as in LIME:
\begin{gather}\label{eq:OC-pointwise}
\mbox{Proposed: }
\max_{\bm{\delta}}\left[\ln \left\{ p(y^t \mid f(\bm{x}^t+\bm{\delta})) \ p(\bm{\delta}) \right\} \right],
\\
\label{eq:obs_model}
p(y\mid f(\bm{x}+\bm{\delta})) =
\mathcal{N}(y \mid f(\bm{x}+\bm{\delta}), \sigma^2(\bm{x})).
\end{gather}
The prior $p(\bm{\delta})$ can be designed to reflect problem-specific constraints such as infeasible regions so that the resultant $\bm{x}+\bm{\delta}$ is a realistic or high probability input. Considering the well-known issue of lasso that in the presence of multiple correlated explanatory variables it tends to pick one at random~\cite{roy2017selection}, we employ $p(\bm{\delta})  \propto \exp\left(
-\frac{1}{2}\lambda \| \bm{\delta} \|_2^2
-  \nu \| \bm{\delta}\|_1
\right)$. $ \sigma^2(\bm{x})$ is the local variance representing the uncertainty of prediction (see Sec.~\ref{subsec:getting_probabilistic_model}), and $\lambda, \nu$ are hyperparameters controlling the sparsity and the overall scale of $\bm{\delta}$ (see Sec.~\ref{subsec:Algo_summary} for typical values). We call $\bm{\delta}$ the \textbf{likelihood compensation} (LC) as it compensates for the loss in likelihood incurred by an anomalous prediction. Note that, unlike LIME, our explainabiliy model is neither linear nor additive, being free from the ``masking effect''~\cite{ESL2} observed in linear XAI models.  

We can naturally extend the point-wise definition of Eq.~\eqref{eq:OC-pointwise} to a collection of test samples. For the Gaussian observation and the elastic net prior, we have the following optimization problem for the LC for $\mathcal{D}_\mathrm{test}$:
\begin{align} \label{eq:OC_problem_definition}
\!
\min_{\bm{\delta}}\left\{\! \frac{1}{N_\mathrm{test}}\!\sum_{t=1}^{N_\mathrm{test}} \!
\frac{\left[ y^t - f(\bm{x}^t + \bm{\delta})\right]^2}{2 \sigma^2_t} \!
+\frac{\lambda}{2} \| \bm{\delta} \|_2^2 + \nu \| \bm{\delta}\|_1
\right\},
\end{align}
where $\sigma^2_t$ is the local variance evaluated at $\bm{x}^t$. \textit{This is the main problem studied in this paper}.

\subsection{Deriving Probabilistic Prediction Model}
\label{subsec:getting_probabilistic_model}

So far we have assumed the predictive distribution $p(y\mid \bm{x})$ is given. Now let us think about how to derive it from the deterministic black-box regression model $y=f(\bm{x})$.

If there are too few test samples, we have no choice but to set $\sigma^2_t$ to a constant using prior knowledge. Otherwise, we can obtain an estimate of $\sigma^2_t$ using a subset of $\mathcal{D}_\mathrm{test}$ in a cross-validation (CV)-like fashion. Let $\mathcal{D}^t_\mathrm{ho}=\{(\bm{x}^{(n)}, y^{(n)}) \mid n=1,\ldots, N_\mathrm{ho} \} \subset \mathcal{D}_\mathrm{test}$ be a held-out data set that does not include the given test sample $(\bm{x}^t,y^t)$. For the observation model Eq.~\eqref{eq:obs_model} and the test sample $\bm{x}^t$, we consider a locally weighted version of maximum likelihood:
\begin{align}\label{eq:ML_for_sigma2}
\max_{\sigma^2}\!
\sum_{n=1}^{N_\mathrm{ho}}\!\! w_n(\bm{x}^t)\left\{
\ln\frac{1}{\sqrt{2\pi\sigma^2}}
 - \frac{(y^{(n)}  - f(\bm{x}^{(n)})  )^2}{2\sigma^2}
\right\},
\end{align}
where $w_n(\bm{x}^t)$ is the similarity between $\bm{x}^t$ and $\bm{x}^{(n)}$. 
A reasonable choice for $w_n$ is the Gaussian kernel:
\begin{align}\label{eq:Gaussian_kernel}
w_n(\bm{x}^t) = \mathcal{N}( \bm{x}^{(n)} \mid \bm{x}^t,\mathrm{diag}(\bm{\eta})),
\end{align}
where $\mathrm{diag}(\bm{\eta})$ is a diagonal matrix whose $i$-th diagonal is given by $\eta_i$, which can be of the same order as the sample variance of $x_i$ evaluated on $\mathcal{D}_\mathrm{ho}$.

The maximizer of Eq.~\eqref{eq:ML_for_sigma2} can be found by differentiating w.r.t.~$\sigma^{-2}$. The solution is given by
\begin{align}\label{eq:sigma2_for_test_samples}
\sigma^2_t = \frac{1}{\sum_{m} w_m(\bm{x}^t)}
\sum_{n=1}^{N_\mathrm{ho}}w_n(\bm{x}^t) \left[y^{(n)}  - f(\bm{x}^{(n)})  \right]^2.
\end{align}
This has to be computed for each $\bm{x}^t \in \mathcal{D}_\mathrm{test}$.

\subsection{Solving the Optimization Problem}
\label{subsec:optimization}

Although seemingly simple, solving the optimization problem~\eqref{eq:OC_problem_definition} is generally challenging. Due to the black-box nature of $f$, we do not have access to the parametric form of $f$, let alone the gradient. In addition, as is the case in deep neural networks, $f$ can be non-smooth (see the red curves in Fig.~\ref{fig:OCillustration.pdf}), which makes numerical estimation of the gradient tricky.

To derive an optimization algorithm, we first note that there are two origins of non-smoothness in the objective function in~\eqref{eq:OC_problem_definition}. One is inherent to $f$ while the other is due to the added $\ell_1$ penalty. To separate them, let us denote the objective function in Eq.~\eqref{eq:OC_problem_definition} as $J(\bm{\delta})+\nu\|\bm{\delta} \|_1$, where $J$ contains the first and second terms. Since we are interested only in a local solution in the vicinity of $\bm{\delta}=\bm{0}$, it is natural to adopt an iterative update algorithm starting from $\bm{\delta}\approx \bm{0}$. Suppose that we have an estimate $\bm{\delta}=\bm{\delta}^\mathrm{old}$ that we wish to update. If we have a reasonable approximation of the gradient in its vicinity, denoted by $\llangle \nabla J(\bm{\delta}^\mathrm{old}) \rrangle$, the next estimate can be found by
\begin{align}\nonumber
\bm{\delta}^\mathrm{new}=
\arg\min_{\bm{\delta}} &\left\{ J(\bm{\delta}^\mathrm{old})+
(\bm{\delta}-\bm{\delta}^\mathrm{old})^\top \llangle\nabla J(\bm{\delta}^\mathrm{old})\rrangle
\right.
\\ \label{eq:prox_Gradient_eq}
&+\left.\frac{1}{2\kappa}\|\bm{\delta}- \bm{\delta}^\mathrm{old} \|_2^2
+ \nu \| \bm{\delta}\|_1
\right\}
\end{align}
in the same spirit of the proximal gradient~\cite{parikh2014proximal}, where $\kappa$ is a hyperparameter representing the learning rate. Notice that the first three terms in the curly bracket correspond to a second-order approximation of $J(\bm{\delta})$ in the vicinity of $\bm{\delta}^\mathrm{old}$. We find the best estimate under this approximation.

The r.h.s.~has an analytic solution. Define $\bm{\phi} \triangleq \bm{\delta}^\mathrm{old} -\kappa \llangle \nabla J(\bm{\delta}^\mathrm{old})\rrangle$. The optimality condition is $ \bm{\delta} - \bm{\phi} + \kappa \nu \ \mathrm{sign}(\bm{\delta})=\bm{0}$. If $\phi_i > \kappa\nu$ holds for the $i$-th dimension, by ${\phi}_i \pm \kappa >0$, we have $\delta_i = \phi_i - \kappa \nu \ \mathrm{sign}(\delta_i) =  \phi_i - \kappa \nu$. Similar arguments straightforwardly verify the following solution:
\begin{align}\label{eq:delta_solution}
    \delta_i =
    \begin{cases}
   \phi_i - \kappa\nu, &\phi_i > \kappa\nu \\
    0, &|\phi_i| \leq \kappa\nu \\
     \phi_i + \kappa\nu, &\phi_i < - \kappa\nu
    \end{cases}.
\end{align}
Performing differentiation, we see that $\bm{\phi}$ is given by
\begin{align}\label{eq:phi_delta}
\bm{\phi} &= (1 - \kappa\lambda)\bm{\delta}^\mathrm{old} +\nonumber\\ &~~~~\kappa
    \frac{1}{N_\mathrm{test}}\sum_{t=1}^{N_\mathrm{test}}
    \left\{
\frac{y^t - f(\bm{x}^t + \bm{\delta}) }{\sigma^2_t }
\right\}
\left\llangle
\frac{\partial f(\bm{x}^t+\bm{\delta})}{\partial \bm{\delta}} \right\rrangle.
\end{align}
Note that $f(\bm{x}^t + \bm{\delta})$ is readily available at any $\bm{\delta}$ without approximation. Here we provide some intuition behind the updating equation~\eqref{eq:phi_delta}. Convergence is achieved when either the deviation $y^t - f$ or the gradient  $\llangle \partial f/\partial \bm{\delta}\rrangle$ vanishes at $\bm{x}^t + \bm{\delta}$. The former and the latter correspond, respectively, to the situations illustrated in Fig.~\ref{fig:OCillustration.pdf}~(a) and~(b). As shown in the figure, $\delta_i$ corresponds to the horizontal deviation along the $x_i$ axis between the test sample and the regression function. If there is no horizontal intersection on the regression surface it seeks the zero gradient point based on a smooth surrogate of the gradient.

To find $\llangle \partial f/\partial \bm{\delta} \rrangle$, a smooth surrogate of the gradient, we propose a simple sampling-based procedure. Specifically, we draw $N_\mathrm{s}$ samples from a local distribution at $\bm{x}^t + \bm{\delta}$ as
\begin{gather}
\bm{x}^{[m]} \sim \mathcal{N}(\cdot \mid \bm{x}^t + \bm{\delta}, \mathrm{diag}(\bm{\eta}) ),
\end{gather}
and fit a linear regression model $f = \beta_0 + \bm{\beta}^\top\bm{x}$ on the populated local data set  $\{ (\bm{x}^{[m]}, f^{[m]} ) \mid m=1,\ldots,N_\mathrm{s}\}$, where $ f^{[m]} =  f( \bm{x}^{[m]}  )$.  Solving the least squares problem, we have
\begin{gather}\label{eq:OLS_for_gradient}
\left\llangle
\frac{\partial f(\bm{x}^t+\bm{\delta})}{\partial \bm{\delta}}
\right\rrangle
 = 
 \bm{\beta}
  = \left[
 \mathsf{\Psi}_\mathrm{s}\mathsf{\Psi}_\mathrm{s}^\top + \varepsilon\mathsf{I}_M
  \right]^{-1}\mathsf{\Psi}_\mathrm{s}\bm{f}_\mathrm{s},
\end{gather}
where $\varepsilon \approx 0$ is a small positive constant added to the diagonals for numerical stability. In Eq.~\eqref{eq:OLS_for_gradient}, we have defined
$
\bm{f}_\mathrm{s} \triangleq [f^{[1]} -\overline{f}, \ldots, f^{[N_\mathrm{s}]} -\overline{f}]^\top
$ and
$
\mathsf{\Psi}_\mathrm{s} \triangleq [ \bm{x}^{[1]} -\overline{\bm{x}}, \ldots, \bm{x}^{[N_\mathrm{s}]} -\overline{\bm{x}} ]
$.
As usual, the population means are defined as $\overline{f} \triangleq \frac{1}{N_\mathrm{s}}\sum_m f^{[m]}$ and $\overline{\bm{x}} \triangleq \frac{1}{N_\mathrm{s}}\sum_m \bm{x}^{[m]}$. 

\subsection{Algorithm Summary}
\label{subsec:Algo_summary}

Algorithm~\ref{algo:OC} summarizes the iterative procedure for finding $\bm{\delta}$. The most important parameter is the $\ell_1$ regularization strength $\nu$, which has to be hand-tuned depending on the business requirements of the application of interest. On the other hand, the $\ell_2$ strength $\lambda$ controls the overall scale of $\bm{\delta}$. It can be fixed to some value between 0 and 1. In our experiments, it was adjusted so its scale is on the same order as LIME's output for consistency.
It is generally recommended to rescale the input variables to have the zero mean and unit variance before starting the iteration (assuming $N_\mathrm{test}\gg 1$), and retrieve the scale factors after convergence. 
For the learning rate $\kappa$, in our experiments, we fixed $\kappa = 0.1$ and shrank it (geometrically) by a factor of 0.98 in every iteration.

\begin{figure}[tb]
\begin{minipage}{0.4\textwidth}
\begin{algorithm}[H]
\caption{Likelihood Compensation}\label{algo:OC}
\label{alg:algorithm}
\textbf{Input}: $f(\bm{x}), \mathcal{D}_\mathrm{test}$.\\
\textbf{Parameters}: $\lambda,\nu,\kappa$.
\begin{algorithmic}[1] 
\FOR{all $\bm{x}^t \in  \mathcal{D}_\mathrm{test}$ }
	\STATE Compute  $\sigma^2_t$ with Eq.~\eqref{eq:sigma2_for_test_samples}.
\ENDFOR
\STATE Randomly initialize $\bm{\delta}\approx \bm{0}$.
\REPEAT
\STATE Set $\bm{g}=\bm{0}$.
\FOR{all $\bm{x}^t \in  \mathcal{D}_\mathrm{test}$ }
	\STATE Compute $\bm{\beta}$ with Eq.~\eqref{eq:OLS_for_gradient}.
	\STATE Update $\bm{g} \leftarrow \bm{g} + \bm{\beta}\frac{y^t - f(\bm{x}^t+\bm{\delta}) }{N_{\mathrm{test}} \sigma^2_t }$.
\ENDFOR
\STATE $\bm{\phi} = (1 - \kappa\lambda)\bm{\delta} + \kappa \bm{g}$.
\STATE Find $\bm{\delta}$ with Eq.~\eqref{eq:delta_solution}.
\UNTIL convergence.
\STATE \textbf{return} $\bm{\delta}$ \\
\end{algorithmic}
\end{algorithm}
\end{minipage}
\end{figure}

In addition to the parameters listed in Algorithm~\ref{algo:OC}, the sampling-based estimation of the gradient Eq.~\eqref{eq:OLS_for_gradient} requires two minor parameters, $N^\mathrm{s},\bm{\eta}$. In the experiment, we fixed $N^\mathrm{s} = 1\,000$ following~\cite{ribeiro2016should} and $\eta_i=1$ for all $i$ after standardization.
The same $\bm{\eta}$ was used for Eq.~\eqref{eq:Gaussian_kernel}.

\section{Experiments}\label{sec:experiments}
We now describe our experimental design and baselines we compare against in the empirical studies that follow.
\paragraph{Evaluation strategy}
Explainability of AI is generally evaluated from three major perspectives~\cite{Costabello2019AAAI_tutorial}: decomposability, simulatability, and algorithmic transparency. In post-hoc explanations of black-box models,  decomposability and simulatability are most important. We thus design our experiments to answer the following questions: a) whether LC can provide extra information on specific anomalous samples beyond the baseline methods (decomposability), b) whether LC can robustly compute the responsibility score under heavy noise (simulatability), and c) whether LC can provide actionable insights in a real-world business scenario (simulatability). Regarding the third question, we validated our approach with feedback from domain experts as opposed to ``crowd sourced'' studies with lay users. In industrial applications, the end-user's interests can be highly specific to particular business needs and the system's inner workings tend to be difficult for non-experts to understand and simulate.

\paragraph{Baselines}
We compare LC with three possible alternatives: (1) $Z$-score and extended versions of (2) Shapley values (SV) and (3) LIME. The $Z$-score is the standard univariate outlier detection method in the unsupervised setting, and that of $x_i^t$ is defined as $(x_i^t-m_i)/\sigma_i$, where $m_i,\sigma_i$ are the mean and the standard deviation of $x_i$ in $\mathcal{D}_\mathrm{test}$, respectively. Shapley values (SV) and LIME are used as a proxy of the prior works~\cite{zhang2019anomaly,giurgiu2019additive,lucic2020does}, which used SV or LIME in certain tasks similar to ours. For fair comparison, we extended these methods to be applied on the deviation $f- y$ instead of $f$ itself, and name them LIME+ and SV+, respectively. We dropped SV+ in the building energy experiment as the training data was not available to compute the null/base values for each variable that SV requires. 
Note that contrastive and counterfactual methods such as \cite{dhurandhar2018explanations,wachter2017counterfactual} are not valid competitors here as they require white-box access to the model and are predominantly used in classification settings.

\begin{figure}[t]
		\centering
		\includegraphics[trim={0.2cm 0.3cm 0cm 0.4cm},clip,width=8cm]{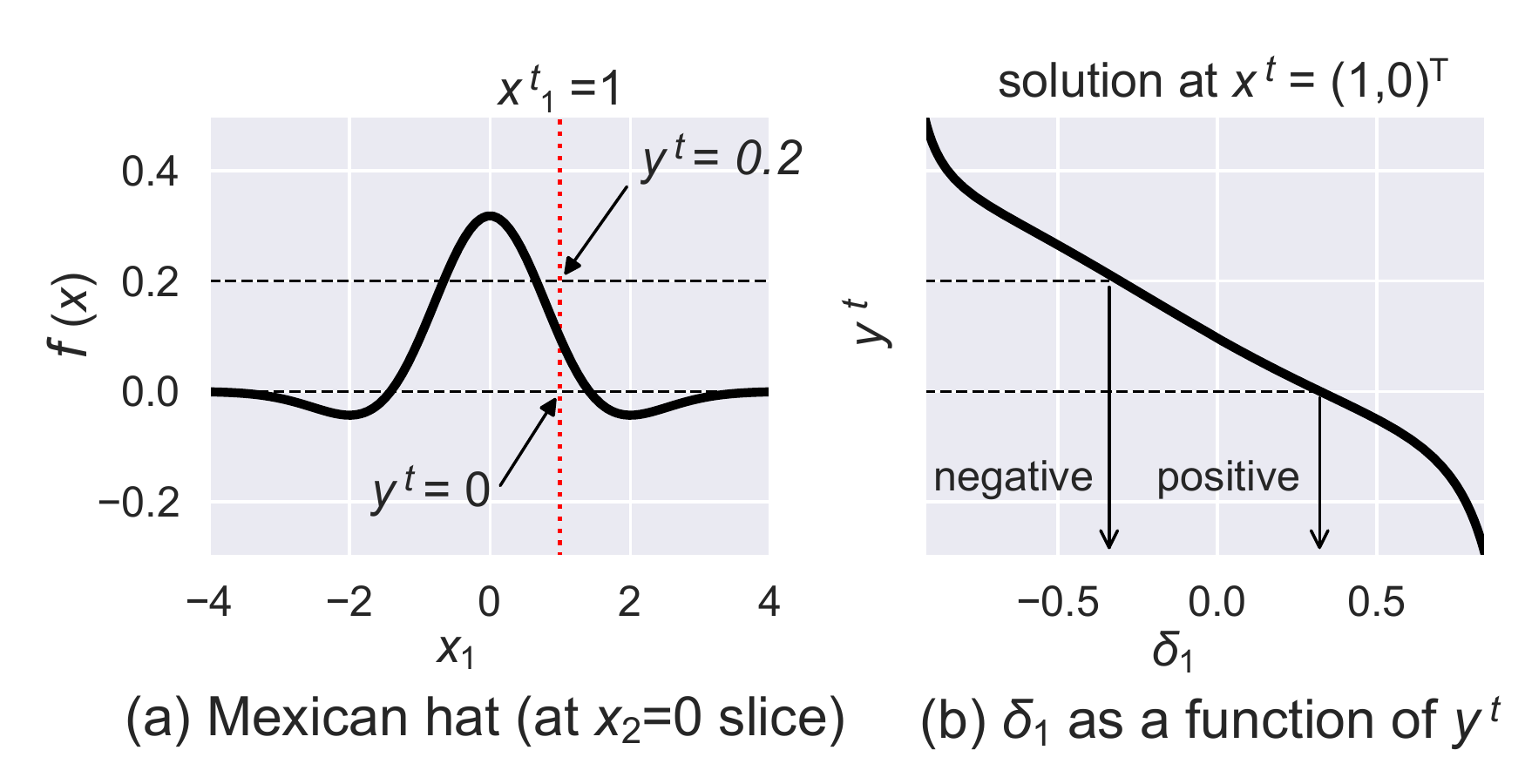}
		\caption{Mexican Hat: (a) The $x_2=0$ slice of $f(\bm{x})$. (b) Computed $\delta_1$ as a function of $y^t$. }
		\label{fig:MexcanHat.pdf}
\end{figure}
	%
\begin{figure}[t]
		\centering
		\includegraphics[trim={0.3cm 0.9cm 0cm 0.4cm},clip,width=8cm]{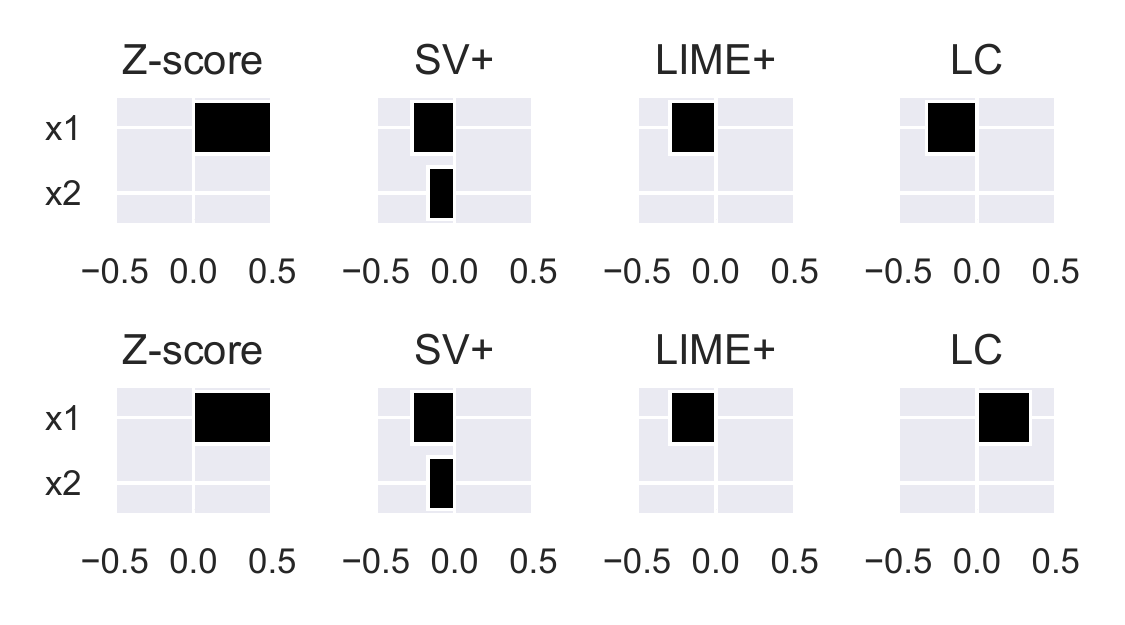}
		\captionof{figure}{Mexican Hat: Comparison of the responsibility scores evaluated at $y^t=0.2$ (upper) and $0$ (lower). }
		\label{fig:MexHat_result}
\end{figure}

\paragraph{Two-Dimensional Mexican Hat} 
One of the major features of LC is its capability to provide explanations relevant to specific anomalous samples. To illustrate this, we used the two-dimensional Mexican Hat for the regression function
$
f(\bm{x})\propto  ( 1- \frac{1}{2}\| \bm{x}\|_2^2)
\exp( - \frac{1}{2}\| \bm{x}\|_2^2 )
$
as shown in Fig.~\ref{fig:MexcanHat.pdf}~(a). Suppose we have obtained a test sample at $\bm{x}^t = (1,0)^\top$. By symmetry, LIME+ has only the $x_1$ component, which can be analytically calculated to be $-0.29$ at this $\bm{x}^t$ when $\nu \rightarrow 0_+$. Similarly, LC has only the $x_1$ component, and is computed through iterative updates with the aid of analytic expression of the gradient. For SV+, we used uniform sampling from $[-4,4]^2$ to evaluate the expectations.  Figure~\ref{fig:MexcanHat.pdf}~(b) shows the calculated values of $\delta_1$ \textit{as a function of $y^t$} with $\nu=0, \lambda=0.01$.

Figure~\ref{fig:MexHat_result} compares $Z$-score, SV+, LIME+, and LC for the two particular values of $y^t$, corresponding to the $f>y^t$ and $f <y^t$ cases. As shown, $Z$-score, SV+, and LIME+ are not able to distinguish between the two cases, 
demonstrating the limited utility in anomaly attribution. 
In contrast, LC's value of $\delta_1$ corresponds to the horizontal distance between the test point and the curve of $f$ as shown in Fig.~\ref{fig:MexcanHat.pdf}. Hence we can think of it as a measure of ``\textit{horizontal deviation},'' as we illustrated earlier in  Fig.~\ref{fig:OCillustration.pdf}.

\begin{figure}[tb]
\centering
\includegraphics[trim={0.cm 0.5cm 0cm 0cm},clip,width=8cm]{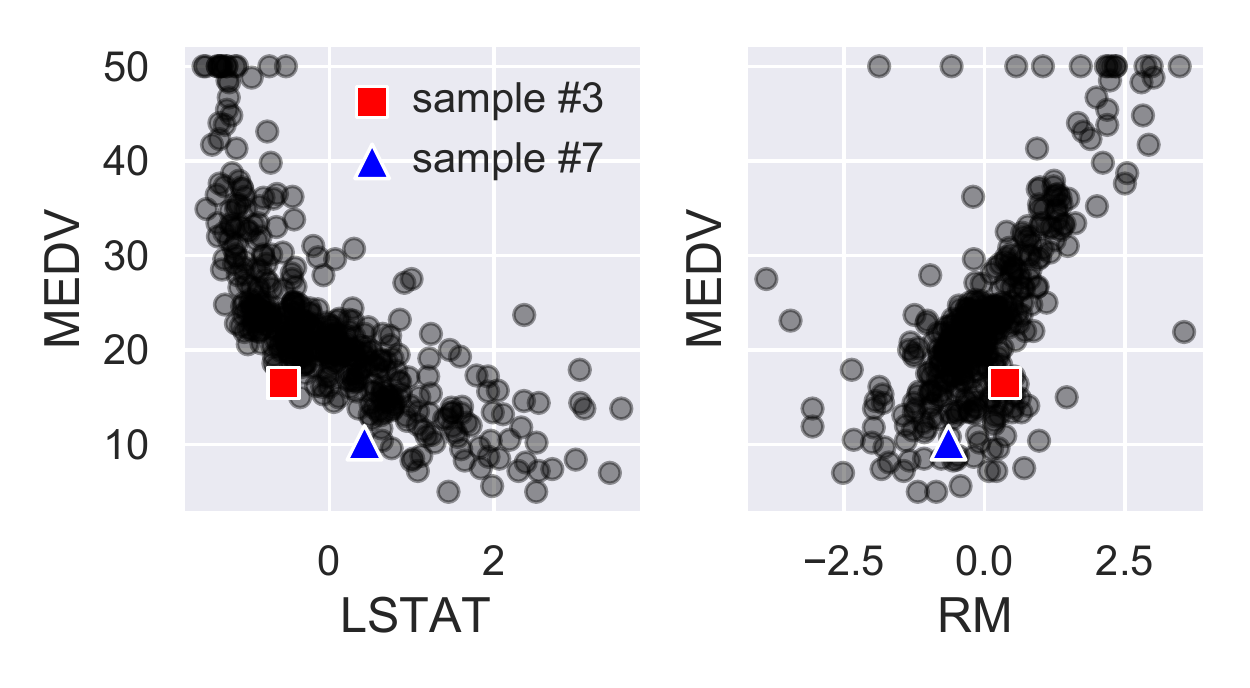}
\vspace{-2mm}
\caption{Boston Housing: Pairwise scatter plot between $y$ (MEDV) and two selected input variables (LSTAT, RM). }
\label{fig:Boston_scatter_LSTATpdf}
\end{figure}

\paragraph{Boston Housing} 
Next we used  Boston Housing data~\cite{BostonHousing} to test the robustness to noise. The task is to predict the median home price (`MEDV') of the districts in Boston with $M=13$ input variables such as the percentage of lower status of the population (`LSTAT') and the average number of rooms (`RM'). As one might expect, the data is very noisy. As an illustrative example, Fig.~\ref{fig:Boston_scatter_LSTATpdf} shows scatter plots between $y$ (MEDV) and two selected input variables (LSTAT, RM), which have the highest correlations with $y$. We held out $20$\% of the data as $\mathcal{D}_\mathrm{test}$ ($N_\mathrm{test}=101$), and trained a random forest on the rest. Then we picked the two top outliers, as highlighted as \#3 and \# 7 in Fig.~\ref{fig:Boston_scatter_LSTATpdf}. These are the two samples with the highest outlier scores of Eq.~\eqref{eq:changeScoreDef}, to which not only LSTAT and RM but also all the other variables contributed.

\begin{figure}[tb]
\centering
\includegraphics[trim={0.cm 0cm 0cm 0cm},clip,width=8cm]{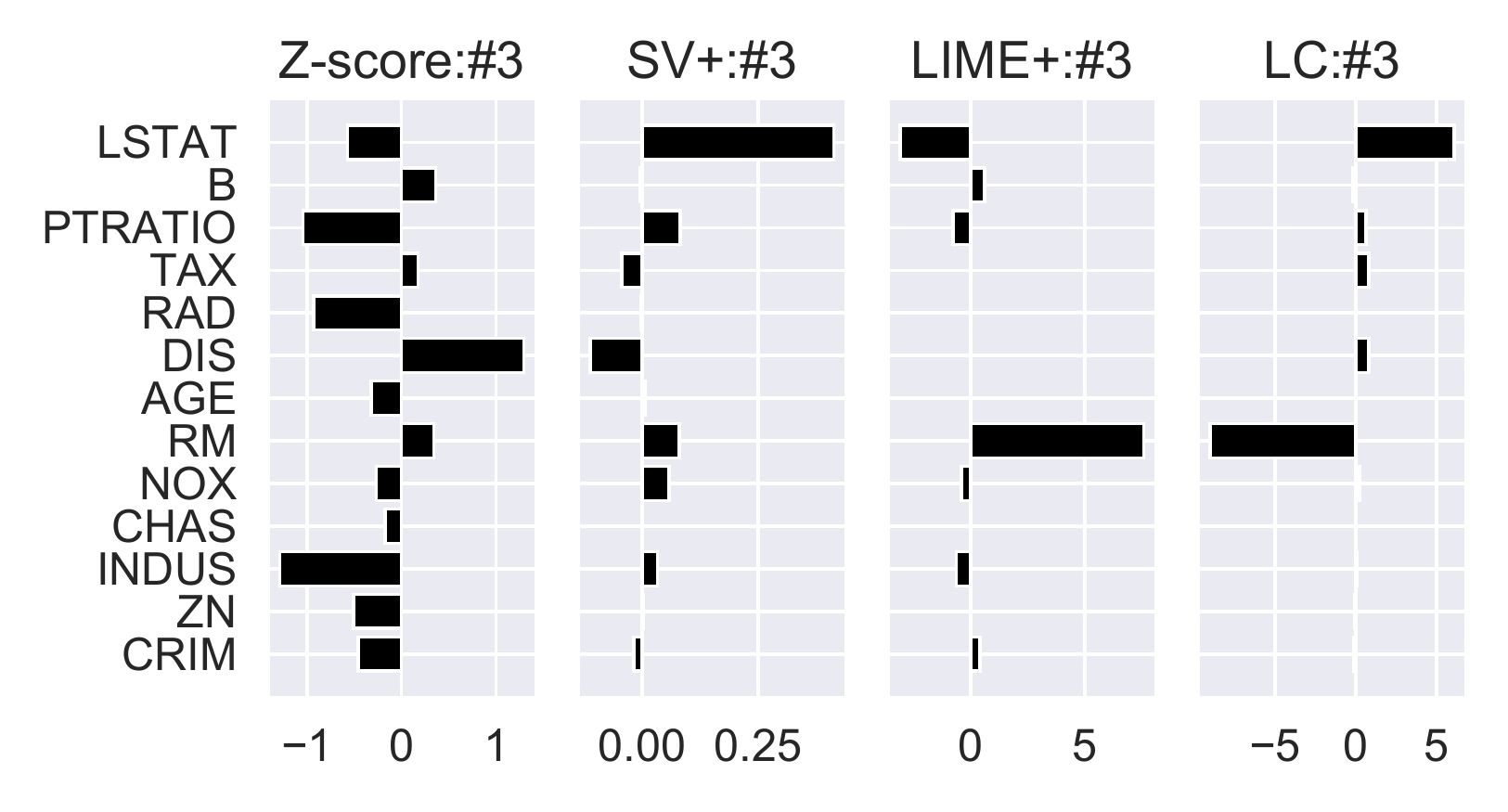}
\includegraphics[trim={0.cm 0.5cm 0cm 0cm},clip,width=8cm]{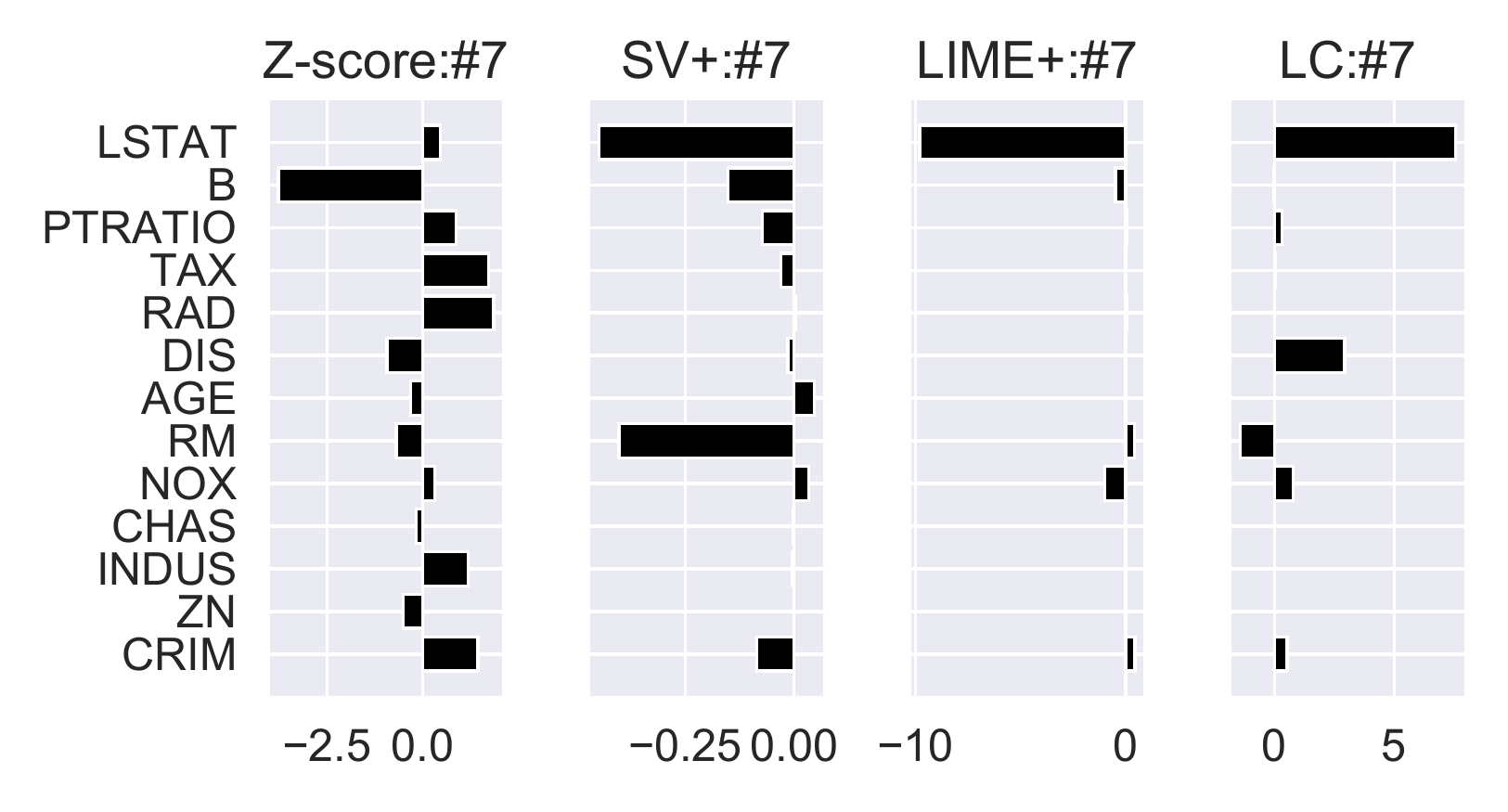}
\caption{Boston Housing: Comparison of the responsibility scores for the top two outliers (\#3 and \#7). }
\label{fig:Boston_3outliers_score_comparison.pdf}
\end{figure}

\begin{figure}[tb]
\begin{center}
\includegraphics[trim={0.cm 0cm 0cm 0cm},clip,width=8cm]{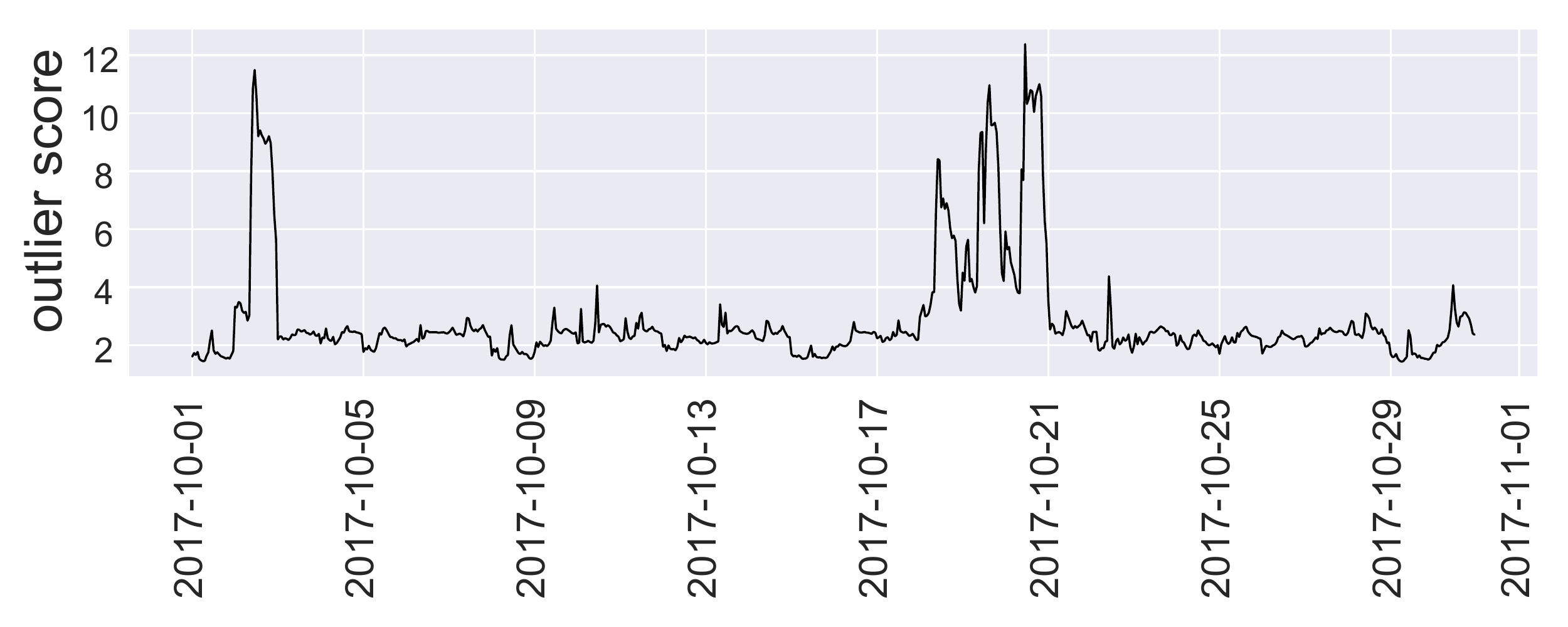}
\end{center}
\vspace{-0.4cm}
\caption{Building Energy: Outlier score computed with Eq.~\eqref{eq:changeScoreDef} for the test data. }
\label{fig:Building_anomalyScore.pdf}
\vspace{-0.4cm}
\end{figure}
\begin{figure*}[t]
\begin{center}
\includegraphics[trim={0.2mm 0.5cm 2.cm 0cm},clip,width=17cm]{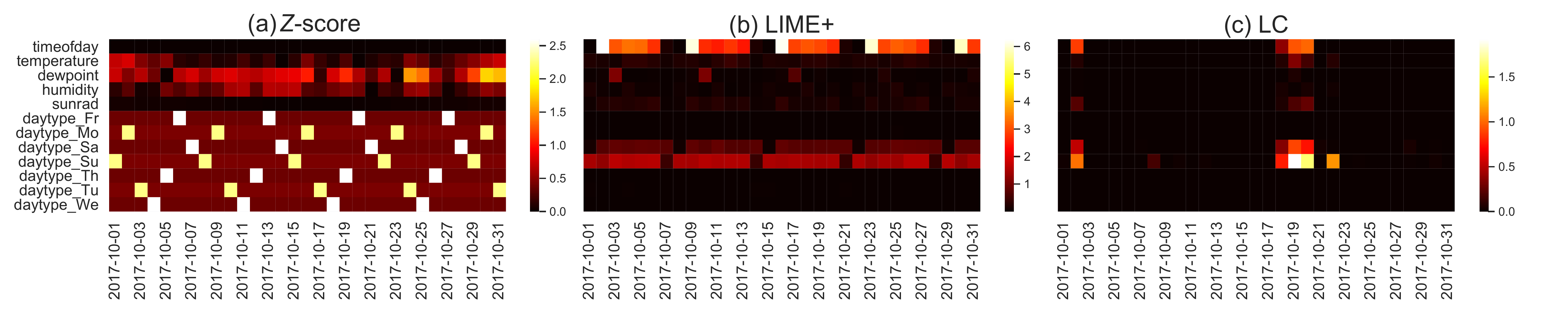}
\end{center}
\vspace{-0.3cm}
\caption{Building Energy: Comparison of the responsibility scores computed for the test data. }
\label{fig:Building_heatmap_GBTree_29_highQ.pdf}
\vspace{-0.3cm}
\end{figure*}

Figure~\ref{fig:Boston_3outliers_score_comparison.pdf} compares the results of LC with the baselines for these outliers. For the $\ell_1$ parameter, we gave $\nu=0.1$ for LC, then chose $\nu=0.005$ for LIME+, so that LIME+ has on average the same number of nonzero elements as LC. The $\ell_2$ parameter $\lambda$ was chosen as $0.5$ for LC and LIME+ to have approximately the same scale. For SV+, all the $2^{M-1}M=53\,248$ combinations were evaluated with the empirical distribution of the training samples, which are actually supposed to be unavailable in our setting, requiring about an hour to finish on a laptop PC (Core i7-8850H) for each test sample, while LC required only several seconds. From the figure, we see that overall SV+, LIME+, and LC are consistent in the sense that most of the weights appear on a few common variables including LSTAT. $Z$-score behaves quite differently, reflecting the fact that it is agnostic to the $y$-$\bm{x}$ relationship. 

For these outliers, LC gave positive and negative scores for LSTAT and RM in Fig.~\ref{fig:Boston_3outliers_score_comparison.pdf}, respectively. Checking the scatter plots in Fig.~\ref{fig:Boston_scatter_LSTATpdf}, we can confirm the LC's characterization as the horizontal deviation that a positive (negative) score means ``a positive (negative) shift will give a better fit.'' In contrast, LIME+ simply indicates whether the local slope is positive or negative, independently how the test samples deviate. In fact, one can mathematically show that LIME+ is invariant to the value of $y$, 
meaning that LIME cannot be a useful tool for instance-specific anomaly attribution. 

In SV+, the situation is more subtle. It does not allow simple interpretations like LC or LIME+. The sign of the scores unpredictably becomes negative or positive, probably due to complicated effects of higher-order correlations. This suggests SV's tendency to be unstable under noise. In fact, our bootstrap analysis (not included for page limitation) 
shows that the SV+ scores are vulnerable to noise; The top three variables with the highest absolute SV+ scores gave a 35.3\% variability relative to the mean. In addition, SV+ needs training data or the true distribution of $\bm{x}$ for Monte Carlo evaluation. $Z$-score, LIME+, and LC do not have such a requirement. 
Along with the prohibitive computational cost, those limitations make it impractical to apply SV+ to real-world system monitoring scenarios of the type presented below.

\paragraph{Real-World Application: Building Energy Management}
Finally, we applied LC to a building administration task. Collaborating with an entity offering building management services and products, we obtained energy consumption data for an office building in India. The total wattage is predicted by a black-box model as a function of weather-related (temperature, humidity, etc.) and time-related variables (time of day, day of week, month, etc.). There are two intended usages of the predictive model. One is near future prediction with short time windows for optimizing HVAC (heating, ventilating, and air conditioning) system control. The other is \textit{retrospective} analysis over the last few months for the purpose of planning long-term improvement of the building facility and its management policies. In the retrospective analysis, it is critical to get clear explanation on unusual events.

At the beginning of the project, we interviewed 10 professionals on what kind of model explainability would be most useful for them. Their top priority capabilities were uncertainty quantification in forecasting and anomaly diagnosis in retrospective analysis. Our choices in the current research reflect these business requirements.

We obtained a one month worth of test data with $M=12$ input variables recorded hourly. We first computed $\sigma^2_t$ according to Eq.~\eqref{eq:sigma2_for_test_samples} in which we leave $(y^t,\bm{x}^t)$ out for each $t$. For each of the test samples, we computed the outlier score by Eq.~\eqref{eq:changeScoreDef} under the Gaussian observation model, which resulted in a few conspicuous anomalies as shown in Fig.~\ref{fig:Building_anomalyScore.pdf}. An important business question was who or what may be responsible for those anomalies.

To obtain insights regarding the detected anomalies, we computed the LC score as shown in Fig.~\ref{fig:Building_heatmap_GBTree_29_highQ.pdf}, where we computed $\bm{\delta}$ each day with $N_\mathrm{test}=24$ in Eq.~\eqref{eq:OC_problem_definition}, and visualized $\|\bm{\delta}\|_2^2$. For the $Z$-score, we visualized the daily mean of the absolute values. For LIME+, we computed regression coefficients for every sample, and visualized the $\ell_2$ norm of their daily mean. We used $(\nu, \lambda)=(0.1,0.5)$, which was determined by the level of sparsity and scale preferred by the domain experts.

As shown in the plot, the LC score clearly highlights a few variables whenever the outlier score is exceptionally high in Fig.~\ref{fig:Building_anomalyScore.pdf}, while the $Z$-score and LIME+ do not provide much information beyond the trivial weekly patterns. The pattern of LIME+ was very stable over $0 <\nu \leq 1$, showing empirical evidence of insensitivity to outliers. As mentioned before, one can mathematically prove this important fact: LIME+ as well as SV+ are invariant to the translation in $f$.
On the other hand, the $Z$-score sensitively captures the variability in the weather-related variables, but it fails to explain the deviations in Fig.~\ref{fig:Building_anomalyScore.pdf}. This is understandable because the $Z$-score does not reflect the relationship between $y$ and $\bm{x}$. The artifact seen in the ``daytype'' variables is due to the one-hot encoding of the day of week.

Finally, with LC, the variables highlighted around October 19 (Thursday) are `timeofday', `daytype\_Sa', and `daytype\_Su', implying that those days had an unusual daily wattage pattern for a weekday and looked more like weekend days. Interestingly, it turned out that the 19th was a national holiday in India and many workers were off on and around that date. Thus we conclude that the anomaly is most likely not due to any faulty building facility, but due to the model limitation caused by the lack of full calendar information. Though simple, such pointed insights made possible by our method were highly appreciated by the professionals.

\section{Conclusions}

We have proposed a new method for model-agnostic explainability in the context of regression-based anomaly attribution. To the best of our knowledge, the proposed method provides the first principled framework for contrastive explainability in regression. The recommended responsibility score Likelihood Compensation is built upon the maximum likelihood principle. This is very different from the objectives used to obtain contrastive explanations in the classification setting. We demonstrated the advantages of the proposed method based on synthetic and real data, as well as on a real-world use-case of building energy management where we sought expert feedback.



\section*{Acknowledgements}
The authors thank Dr.~Kaoutar El Maghraoui for her support and technical suggestions. T.I.~is partially supported by the Department of Energy National Energy Technology Laboratory under Award Number DE-OE0000911. A part of this report was prepared as an account of work sponsored by an agency of the United States Government. Neither the United States Government nor any agency thereof, nor any of their employees, makes any warranty, express or implied, or assumes any legal liability or responsibility for the accuracy, completeness, or usefulness of any information, apparatus, product, or process disclosed, or represents that its use would not infringe privately owned rights. Reference herein to any specific commercial product, process, or service by trade name, trademark, manufacturer, or otherwise does not necessarily constitute or imply its endorsement, recommendation, or favoring by the United States Government or any agency thereof. The views and opinions of authors expressed herein do not necessarily state or reflect those of the United States Government or any agency thereof.

\bibliography{ide_et_al}

\begin{thebibliography}{28}
\providecommand{\natexlab}[1]{#1}
\providecommand{\url}[1]{\texttt{#1}}
\providecommand{\urlprefix}{URL }
\expandafter\ifx\csname urlstyle\endcsname\relax
  \providecommand{\doi}[1]{doi:\discretionary{}{}{}#1}\else
  \providecommand{\doi}{doi:\discretionary{}{}{}\begingroup
  \urlstyle{rm}\Url}\fi

\bibitem[{Amarasinghe, Kenney, and Manic(2018)}]{amarasinghe2018toward}
Amarasinghe, K.; Kenney, K.; and Manic, M. 2018.
\newblock Toward explainable deep neural network based anomaly detection.
\newblock In \emph{Proc. Intl. Conf. Human System Interaction (HSI)}, 311--317.
  IEEE.

\bibitem[{Bach et~al.(2015)Bach, Binder, Montavon, Klauschen, M{\"u}ller, and
  Samek}]{bach2015pixel}
Bach, S.; Binder, A.; Montavon, G.; Klauschen, F.; M{\"u}ller, K.-R.; and
  Samek, W. 2015.
\newblock On pixel-wise explanations for non-linear classifier decisions by
  layer-wise relevance propagation.
\newblock \emph{PloS one} 10(7): e0130140.

\bibitem[{Belsley(1980)}]{BostonHousing}
Belsley, K. .~W. 1980.
\newblock \emph{Regression diagnostics: Identifying Influential Data and
  Sources of Collinearity}.
\newblock Wiley.

\bibitem[{Casalicchio, Molnar, and Bischl(2018)}]{casalicchio2018visualizing}
Casalicchio, G.; Molnar, C.; and Bischl, B. 2018.
\newblock Visualizing the feature importance for black box models.
\newblock In \emph{Proc. Joint European Conference on Machine Learning and
  Knowledge Discovery in Databases}, 655--670. Springer.

\bibitem[{Chandola, Banerjee, and Kumar(2009)}]{Chandola09AnomalySurvey}
Chandola, V.; Banerjee, A.; and Kumar, V. 2009.
\newblock Anomaly Detection: A Survey.
\newblock \emph{ACM Computing Survey} 41(3): 1--58.

\bibitem[{Costabello et~al.(2019)Costabello, Giannotti, Guidotti, Hitzler,
  L\'ecu\'e, Minervini, and Sarker}]{Costabello2019AAAI_tutorial}
Costabello, L.; Giannotti, F.; Guidotti, R.; Hitzler, P.; L\'ecu\'e, F.;
  Minervini, P.; and Sarker, K. 2019.
\newblock On Explainable {AI}: From Theory to Motivation, Applications and
  Limitations.
\newblock In \emph{Tutorial, AAAI Conference on Artificial Intelligence}.

\bibitem[{Dhurandhar et~al.(2018)Dhurandhar, Chen, Luss, Tu, Ting, Shanmugam,
  and Das}]{dhurandhar2018explanations}
Dhurandhar, A.; Chen, P.-Y.; Luss, R.; Tu, C.-C.; Ting, P.; Shanmugam, K.; and
  Das, P. 2018.
\newblock Explanations based on the missing: Towards contrastive explanations
  with pertinent negatives.
\newblock In \emph{Advances in Neural Information Processing Systems},
  592--603.

\bibitem[{Fong and Vedaldi(2017)}]{fong2017interpretable}
Fong, R.~C.; and Vedaldi, A. 2017.
\newblock Interpretable explanations of black boxes by meaningful perturbation.
\newblock In \emph{Proc. IEEE Intl. Conf. Computer Vision}, 3429--3437.

\bibitem[{Gade et~al.(2020)Gade, Geyik, Kenthapadi, Mithal, and
  Taly}]{gade2020explainable}
Gade, K.; Geyik, S.; Kenthapadi, K.; Mithal, V.; and Taly, A. 2020.
\newblock Explainable AI in Industry: Practical Challenges and Lessons Learned.
\newblock In \emph{Companion Proceedings of the Web Conference 2020}, 303--304.

\bibitem[{Giurgiu and Schumann(2019)}]{giurgiu2019additive}
Giurgiu, I.; and Schumann, A. 2019.
\newblock Additive Explanations for Anomalies Detected from Multivariate
  Temporal Data.
\newblock In \emph{Proc. Intl. Conf. Information and Knowledge Management},
  2245--2248. ACM.

\bibitem[{Hastie, Tibshirani, and Friedman(2009)}]{ESL2}
Hastie, T.; Tibshirani, R.; and Friedman, J. 2009.
\newblock \emph{{ The Elements of Statistical Learning: Data Mining, Inference,
  and Prediction}}.
\newblock Springer, 2 edition.

\bibitem[{Langone, Cuzzocrea, and Skantzos(2020)}]{langone2020interpretable}
Langone, R.; Cuzzocrea, A.; and Skantzos, N. 2020.
\newblock Interpretable Anomaly Prediction: Predicting anomalous behavior in
  industry 4.0 settings via regularized logistic regression tools.
\newblock \emph{Data \& Knowledge Engineering} 101850.

\bibitem[{Lucic, Haned, and de~Rijke(2020)}]{lucic2020does}
Lucic, A.; Haned, H.; and de~Rijke, M. 2020.
\newblock Why does my model fail? contrastive local explanations for retail
  forecasting.
\newblock In \emph{Proceedings of the 2020 Conference on Fairness,
  Accountability, and Transparency}, 90--98.

\bibitem[{Lundberg and Lee(2017)}]{lundberg2017unified}
Lundberg, S.~M.; and Lee, S.-I. 2017.
\newblock A unified approach to interpreting model predictions.
\newblock In \emph{Advances in Neural Information Processing Systems},
  4765--4774.

\bibitem[{Molnar(2019)}]{molnar2019interpretable}
Molnar, C. 2019.
\newblock \emph{Interpretable machine learning}.
\newblock Lulu.

\bibitem[{Parikh, Boyd et~al.(2014)}]{parikh2014proximal}
Parikh, N.; Boyd, S.; et~al. 2014.
\newblock Proximal algorithms.
\newblock \emph{Foundations and Trends in Optimization} 1(3): 127--239.

\bibitem[{Ribeiro, Singh, and Guestrin(2016)}]{ribeiro2016should}
Ribeiro, M.~T.; Singh, S.; and Guestrin, C. 2016.
\newblock Why should {I} trust you?: Explaining the predictions of any
  classifier.
\newblock In \emph{Proc. ACM SIGKDD Intl. Conf. Knowledge Discovery and Data
  Mining}, 1135--1144. ACM.

\bibitem[{Ribeiro, Singh, and Guestrin(2018)}]{ribeiro2018anchors}
Ribeiro, M.~T.; Singh, S.; and Guestrin, C. 2018.
\newblock Anchors: High-precision model-agnostic explanations.
\newblock In \emph{Proc. AAAI Conference on Artificial Intelligence}.

\bibitem[{Roy, Chakraborty et~al.(2017)}]{roy2017selection}
Roy, V.; Chakraborty, S.; et~al. 2017.
\newblock Selection of tuning parameters, solution paths and standard errors
  for {Bayesian} lassos.
\newblock \emph{Bayesian Analysis} 12(3): 753--778.

\bibitem[{Samek et~al.(2019)Samek, Montavon, Vedaldi, Hansen, and
  M{\"u}ller}]{samek2019explainable}
Samek, W.; Montavon, G.; Vedaldi, A.; Hansen, L.~K.; and M{\"u}ller, K.-R.
  2019.
\newblock \emph{Explainable AI: interpreting, explaining and visualizing deep
  learning}, volume 11700.
\newblock Springer Nature.

\bibitem[{Sipple(2020)}]{SippleICML2020}
Sipple, J. 2020.
\newblock Interpretable, Multidimensional, Multimodal Anomaly Detection with
  Negative Sampling for Detection of Device Failure.
\newblock In \emph{Proc. Intl. Conf. Machine Learning}, 9016--9025.

\bibitem[{{\v{S}}trumbelj and Kononenko(2010)}]{kononenko2010efficient}
{\v{S}}trumbelj, E.; and Kononenko, I. 2010.
\newblock An efficient explanation of individual classifications using game
  theory.
\newblock \emph{Journal of Machine Learning Research} 11(Jan): 1--18.

\bibitem[{{\v{S}}trumbelj and Kononenko(2014)}]{vstrumbelj2014explaining}
{\v{S}}trumbelj, E.; and Kononenko, I. 2014.
\newblock Explaining prediction models and individual predictions with feature
  contributions.
\newblock \emph{Knowledge and information systems} 41(3): 647--665.

\bibitem[{Sundararajan, Taly, and Yan(2017)}]{sundararajan2017axiomatic}
Sundararajan, M.; Taly, A.; and Yan, Q. 2017.
\newblock Axiomatic attribution for deep networks.
\newblock In \emph{Proc. Intl. Conf. Machine Learning}, 3319--3328.

\bibitem[{Tao et~al.(2018)Tao, Cheng, Qi, Zhang, Zhang, and
  Sui}]{tao2018digital}
Tao, F.; Cheng, J.; Qi, Q.; Zhang, M.; Zhang, H.; and Sui, F. 2018.
\newblock Digital twin-driven product design, manufacturing and service with
  big data.
\newblock \emph{The International Journal of Advanced Manufacturing Technology}
  94(9-12): 3563--3576.

\bibitem[{Wachter, Mittelstadt, and Russell(2017)}]{wachter2017counterfactual}
Wachter, S.; Mittelstadt, B.; and Russell, C. 2017.
\newblock Counterfactual explanations without opening the black box: Automated
  decisions and the {GDPR}.
\newblock \emph{Harvard Journal of Law \& Technology} 31: 841.

\bibitem[{Zemicheal and Dietterich(2019)}]{Zemicheal19COMPASS}
Zemicheal, T.; and Dietterich, T.~G. 2019.
\newblock Anomaly detection in the presence of missing values for weather data
  quality control.
\newblock In \emph{Proc. ACM SIGCAS Conf. Computing and Sustainable Societies},
  65--73.

\bibitem[{Zhang et~al.(2019)Zhang, Marwah, Lee, Arlitt, and
  Goldwasser}]{zhang2019anomaly}
Zhang, X.; Marwah, M.; Lee, I.-t.; Arlitt, M.; and Goldwasser, D. 2019.
\newblock {ACE}--{An} Anomaly Contribution Explainer for Cyber-Security
  Applications.
\newblock In \emph{Proc. IEEE Intl. Conf. Big Data}, 1991--2000.

\end{thebibliography}

\end{document}